\documentclass[runningheads]{llncs}

\usepackage{graphicx}
\usepackage{hyperref}
\usepackage{amssymb}
\usepackage{amsmath}

\DeclareMathOperator*{\argmin}{argmin}
\newcommand{\dg}{$^\circ$}

\usepackage{color}

\begin{document}


\title{Spatial regularisation for improved accuracy and interpretability in keypoint-based registration}

\titlerunning{Spatial regularisation for keypoint-based registration}

\author{
Benjamin Billot \inst{1} \and
Ramya Muthukrishnan\inst{2} \and
Esra Abaci-Turk\inst{3} \and \\
P. Ellen Grant\inst{3} \and
Nicholas Ayache \inst{1} \and
Herv\'e Delingette \inst{1} \and
Polina Golland\inst{2}
}
\institute{
Inria, Epione team, Sophia-Antipolis, France 
\and
Massachusetts Institute of Technology, Cambridge, USA
\and
Boston Children’s Hospital and Harvard Medical School, Boston, USA \\
\email{benjamin.billot@inria.fr}}
\authorrunning{B. Billot et al.}
\maketitle

\begin{abstract} 

Unsupervised registration strategies bypass requirements in ground truth transforms or segmentations by optimising similarity metrics between fixed and moved volumes. Among these methods, a recent subclass of approaches based on unsupervised keypoint detection stand out as very promising for interpretability. Specifically, these methods train a network to predict feature maps for fixed and moving images, from which explainable centres of mass are computed to obtain point clouds, that are then aligned in closed-form. However, the features returned by the network often yield spatially diffuse patterns that are hard to interpret, thus undermining the purpose of keypoint-based registration. Here, we propose a three-fold loss to regularise the spatial distribution of the features. First, we use the KL divergence to model features as point spread functions that we interpret as probabilistic keypoints. Then, we sharpen the spatial distributions of these features to increase the precision of the detected landmarks. Finally, we introduce a new repulsive loss across keypoints to encourage spatial diversity. Overall, our loss considerably improves the interpretability of the features, which now correspond to precise and anatomically meaningful landmarks. We demonstrate our three-fold loss in foetal rigid motion tracking and brain MRI affine registration tasks, where it not only outperforms state-of-the-art unsupervised strategies, but also bridges the gap with state-of-the-art supervised methods. 
Our code is available at \href{https://github.com/BenBillot/spatial_regularisation}{https://github.com/BenBillot/spatial\_regularisation}.


\end{abstract}

\section{Introduction}

Registration is paramount in medical image analysis for tasks such as longitudinal studies~\cite{holland_longitudinal_2011} or atlas building~\cite{joshi_atlas_2004}. Modern strategies are dominated by learning-based methods, which enable fast inference compared to classical optimisation frameworks, for example ANTs~\cite{avants_symmetric_2008} and NiftyReg~\cite{modat_niftyreg_2014}. Deep learning approaches have first been optimised using supervision provided as ground truths transforms (either obtained by construction~\cite{eppenhof_pulmonary_2018,uzunova_training_2017} or from classical optimisation frameworks~\cite{fan_birnet_2019,salehi_3dposenet_2019}), or segmentations~\cite{hu_weakly_2018,lee_image_2019}. Learning-based registration methods have been shifting towards unsupervised registration methods, where networks are trained using similarity metrics between fixed and moved volumes~\cite{balakrishnan_voxelmorph_2019,dalca_unsupervised_2019,devos_dlir_2019}, thus bypassing the need for supervision.

\begin{figure}[t]
\includegraphics[width=\textwidth]{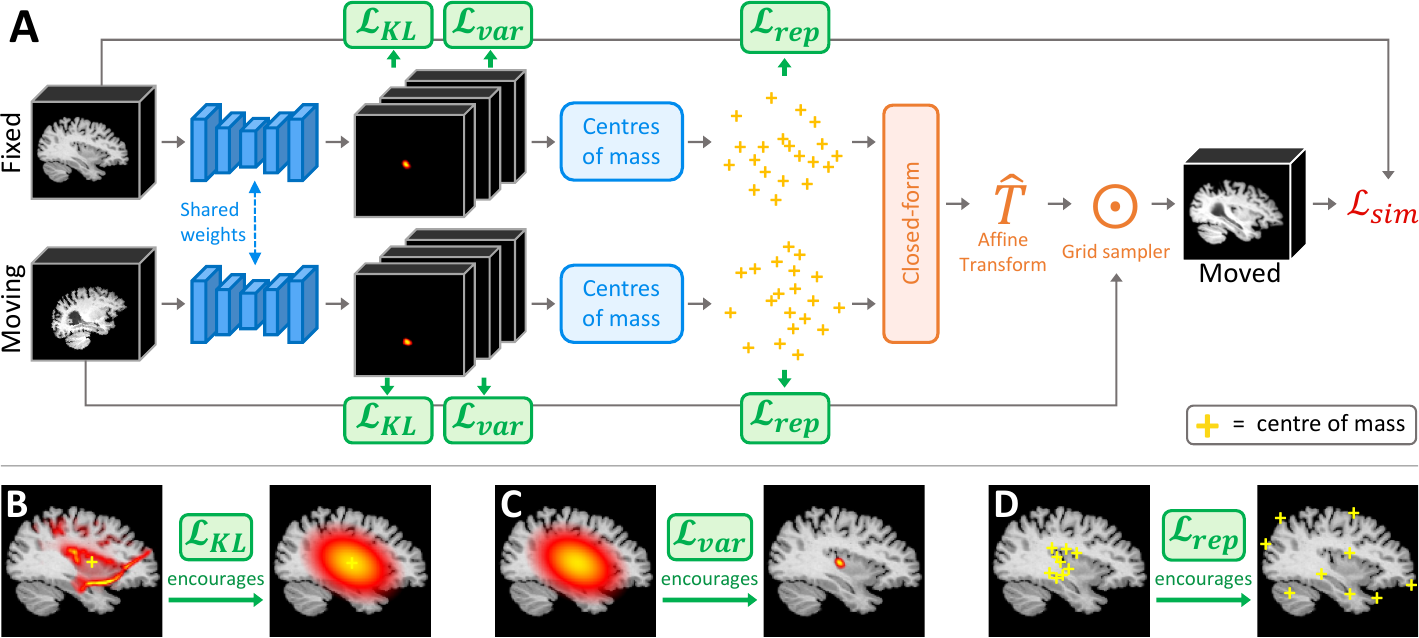}
\caption{(A) Registration by unsupervised keypoint detection. Beyond the classical similarity loss between fixed and moved volumes (leading to features comparable to B,left), we propose a three-fold spatial regularisation, whose effects on the features/keypoints are depicted in (B-D): interpretability ($\mathcal{L}_{KL}$,), precision ($\mathcal{L}_{var}$), and diversity ($\mathcal{L}_{rep}$).} 
\label{fig:overview}
\end{figure}

Meanwhile, the interpretability of learning-based approaches has been an active area of research~\cite{chen_survey_2025}. Following a rich literature of landmark registration~\cite{chui_point_2003,pennec_landmark_1997,tuytelaars_landmarks_2008}, a recent class of methods based on unsupervised keypoint detection has emerged as particularly promising for interpretability~\cite{moyer_equivariant_2021,evan_keymorph_2022}, where warps are estimated by: (\textit{i})~passing fixed and moving volumes through a siamese network, either implemented as a regular~\cite{wang_keymorph_2023} or equivariant~\cite{billot_equitrack_2024} convolutional neural network; (\textit{ii})~computing keypoints as the centres of mass of all output features;(\textit{iii})~aligning the resulting point clouds in closed-form. Conceptually, these methods are explainable through their learnt keypoints, whose correspondence across volumes can be verified. However, in practice, these networks often present irregular and spatially diffuse features that remain hard to interpret (Fig.~\ref{fig:overview}.B,left). Hence, centres of mass are poor summaries of such features, sometimes making keypoints difficult to reproduce across volumes. Moreover, computing centres of mass over these diffuse features results in many keypoints landing at the image centre~\cite{wang_keymorph_2023}, which leads to suboptimal transform estimates.

Here, we propose to tackle these problems by regularising the spatial distribution of the network's features. Spatial regularisation has a long history in registration, mainly in the elastic case~\cite{chen_survey_2025}, where unrealistic deformations are avoided by using diffeomorphisms~\cite{avants_symmetric_2008,lorenzi_lcc_2013} or penalising deformation fields with high bending energy~\cite{rueckert_bsplines_1999}. However, these strategies do not apply to our keypoint-like features, which are very different from deformation fields. Spatial regularisation also has been of high interest in segmentation, to ensure anatomical plausibility of the returned masks~\cite{liu_anatomy_2021}. Closer to our work are two spatial regularisation losses for \textit{supervised} keypoint detection. First, one can encourage the network's output features to resemble point spread functions by optimising a Kullback-Leibler (KL) divergence with isotropic Gaussians of fixed variance~\cite{nibali_dnst_2018}. To avoid finding an optimal dataset-dependent variance, another method directly minimises the second order moment (i.e., the spatial variance) of the features~\cite{fang_var_2023}. Yet, this latter approach remains suboptimal for unsupervised keypoint detection, as it does not explicitly penalise spatial distributions that are not uni-modal Gaussians.

Building on these ideas, we propose a new three-fold loss function to improve the accuracy and interpretability of registration methods based on unsupervised keypoint detection. First, we regularise the network's features by minimising the KL divergence with Gaussians of corresponding means and covariance matrices, which we sharpen with a second term penalising spatial variance. As a result, the network produces sharp point spread functions interpretable as precise keypoints. Compared to~\cite{nibali_dnst_2018}, decoupling the KL loss from the variance regularisation enables the network to be more flexible in its choice of keypoints, rather than having to find landmarks that satisfy an arbitrary target variance. Finally, we further improve performance by penalising redundancy in the detected keypoints with a new repulsive loss that encourages spatial diversity. We demonstrate our loss on two unsupervised keypoint detection methods for rigid motion tracking in foetal time series and affine registration in brain MRI. Our method outperforms other unsupervised strategies and obtains comparable results to state-of-the-art supervised registration methods, while considerably increasing interpretability.

\section{Methods}

\subsection{Registration by unsupervised keypoint detection}

Our method builds on registration by unsupervised keypoint detection, which we summarise below and illustrate in Fig.~\ref{fig:overview}.A. While this class of methods also applies to non-linear registration~\cite{wang_keymorph_2023}, we focus here on affine deformations.

Let $I^m$ and $I^f$ be moving and fixed volumes defined on a finite discretised grid $\Omega \subset \mathbb{R}^3$. We train a convolutional neural network (CNN) $\Phi_\theta$, parametrised by $\theta$, to obtain an affine transform $\hat{T}$ that maximises the alignment of $I^m$ with $I^f$. Inspired by classical optimisation frameworks, unsupervised learning strategies solve this problem by optimising a similarity loss $\mathcal{L}_{sim}$ between the fixed and moved volumes: $\theta^* = \argmin_\theta \; \mathbb{E}_{\{I^f,I^m\}}\left[ \mathcal{L}_{sim} \left( I^f, \, \Phi_\theta(I^f,I^m) \circ I^m \right) \right]$. Most methods~\cite{devos_dlir_2019,salehi_3dposenet_2019} use $\Phi_\theta$ to directly regress the parameters of the affine transform: $\hat{T} = \Phi_\theta(I^f,I^m)$, where $\hat{T}$ is a 4$\!\times\!$4 matrix in homogeneous coordinates. However, this approach lacks interpretability and is fragile against large misalignments.

Instead of direct regression, registration by unsupervised keypoint detection proposes to train $\Phi_\theta$ to extract two 3D point clouds, each with $K$ points, from $I^f$ and $I^m$: $\mu^f=\{\mu^f_k\}_{k=1}^K$ and $\mu^m=\{\mu^m_k\}_{k=1}^K$, respectively. This is achieved by predicting $K$ feature maps for each volume and computing their centres of mass. Crucially, we assume that $\Phi_\theta$ learns to predict keypoints that correspond to matching anatomical landmarks across volumes and poses, such that $\mu^f = \hat{T} \mu^m$. Thus, $\hat{T}$ can now be estimated from the obtained point clouds by minimising the distance between corresponding pairs of points: $\hat{T} = \argmin_T \Sigma_{k=1}^K ||\mu_k^f - T\mu_k^m||^2$, which is done in differentiable closed-form~\cite{horn_svd_1987,kabsch_svd_1976,evan_keymorph_2022}.

While this framework is trained with the same loss $\mathcal{L}_{sim}$ as regression-based approaches (Fig~\ref{fig:overview}.A), its results are more interpretable since we have access to the keypoints that the network has identified as reliable to robustly estimate $\hat{T}$. Implementations of unsupervised keypoint-based methods mostly differ by the choice of the network architecture and closed-form algorithm: rigid transforms are tackled with SE(3)-equivariant networks~\cite{moyer_equivariant_2021,billot_equitrack_2024} and SVD-based closed-form solutions~\cite{kabsch_svd_1976,horn_svd_1987,chatrasingh_svd-scaling_2023}, whereas affine registration is performed using regular CNNs along with Moore-Penrose pseudo-inverse fitting of point sets~\cite{evan_keymorph_2022,wang_keymorph_2023}. 

However, in all cases, the feature maps extracted by the network often present irregular and diffuse patterns (Fig.~\ref{fig:overview}.B,left), which is problematic for three reasons. First, these patterns are hard to interpret, thus undermining the very purpose of keypoint-based registration. Second, the feature maps are difficult to reproduce across volumes, which introduces noise in the extracted keypoints. Finally, spatially averaging these diffuse feature maps leads to centres of mass that are clustered at the image centre, which is suboptimal for transform estimation.

\subsection{Proposed spatial regularisation loss}
\label{sec:proposed losses}

We address the aforementioned problems by introducing a new principled three-fold loss to regularise the spatial distribution of the network's features. Here, we describe each component separately and provide illustrations in Fig.~\ref{fig:overview}.B-D.

\subsubsection{Interpretable feature maps.}
We first seek to improve the interpretability of the feature maps extracted by $\Phi_\theta$, because their current spatial distributions are not only difficult to interpret, but also poorly justify using centres of mass to describe them. Thus, we propose to regularise the network's features such that they yield spatial distributions that are densely concentrated around their centres of mass. For each feature map, this forces the network to focus on one specific landmark, which can be interpreted as the detected keypoint (Fig.~\ref{fig:overview}.B). 

Specifically, we regularise each feature map separately by computing the KL divergence with a spatial Gaussian distribution of the same mean and covariance matrix. Let $F_k(X)$ be the $k^{th}$ feature map returned by $\Phi_\theta$ defined on the voxels $X\in \Omega$. After using a softmax to make $F_k$ positive and normalised (i.e., $\sum_XF_k(X)=1$), we compute its centre of mass $\mu_k\!=\!\sum_XXF_k(X)$ and $3\!\times\!3$ covariance matrix $\Sigma_k\!=\! \sum_XF_k(X)(X-\mu_k)(X-\mu_k)^T$. If $|\Omega|$ is the cardinality of $\Omega$, and $\mathcal{N}(\cdot)$ is the normal distribution, our first regularisation term $\mathcal{L}_{KL}$ is:
\begin{equation}
    \mathcal{L}_{KL} =\frac{1}{K} \frac{1}{|\Omega|}  \sum_{k=1}^K \sum_{X\in\Omega}  F_k(X) \left[ \log{F_k(X)} - \log{\mathcal{N}(X|\mu_k, \Sigma_k)}\right].
\end{equation}
Here $\Sigma_k$ is a covariance matrix rather than a scalar variance to account for~potentially anisotropic landmarks (e.g., keypoints at the boundary between regions).

\subsubsection{Precise landmark detection.} 
After regularising the feature maps to become interpretable point spread functions, we add another loss $\mathcal{L}_{var}$ aiming at reducing the spatial variance of the obtained features. The objective is to obtain keypoints anchored to \textit{precise} anatomical landmarks (e.g., boundary between regions, apex of a structure), rather than large areas (Fig.~\ref{fig:overview}.C). Extending~\cite{fang_var_2023}, we regularise the covariance matrix of each feature map with the Frobenius norm $||\cdot||_F$:
\begin{equation}
    \mathcal{L}_{var} = \frac{1}{K} \sum_{k=1}^K ||\Sigma_k||_F = \frac{1}{K} \sum_{k=1}^K \sqrt{Tr(\Sigma_k \Sigma_k^T)} = \frac{1}{K} \sum_{k=1}^K \left(\frac{1}{9}\sum_{i=1}^9 \Sigma_{k,i}^2\right)^{1/2},
\end{equation}
where $Tr$ is the trace, and $i$ indexes the elements of $\Sigma_k$. By explicitly regularising all values in $\Sigma_k$, this loss is slightly more efficient than~\cite{fang_var_2023}, which only regularises diagonal elements of $\Sigma_k$ (i.e., implicit regularisation of non-diagonal values).

\subsubsection{Diversity of keypoints.} 
Although we now obtain interpretable and precise keypoints, we still observe that many of them are redundant and closely clustered near the image centre (Fig.~\ref{fig:overview}.D). \cite{evan_keymorph_2022,wang_keymorph_2023} attempt to address this issue by spreading the network's attention across the image space, which is done by pretraining $\Phi_\theta$ to consistently detect random points in a single input volume that undergoes online intensity and spatial augmentations. We show below that this strategy yields only slight improvements and propose instead to explicitly maximise the distances between every pair of keypoints $\mu_k$ and $\mu_{k'}$ ($k \in \{1,...,K\}, k'>k$). For numerical stability, we take the sigmoid of distances (so that they are bounded by 1), and minimise the negative logarithm, which results in $\mathcal{L}_{rep}$:
\begin{equation}
    \mathcal{L}_{rep} = \frac{2}{K(K-1)} \sum_{k'>k} - \log{ \left(\frac{1}{1 + \exp{[-\frac{||\mu_k - \mu_{k'}||_2}{\tau}]}} \right)},
\end{equation}
where $\tau>0$ is a hyperparameter controlling the slope of the sigmoid. We emphasise that this loss has a strong connection to the traditional contrastive loss~\cite{wang_understanding_2020}, from which we differ in three ways: (\textit{i})~we replace the cosine similarity with the Euclidean distance, (\textit{ii})~we only consider negative pairs (all points must be as far away as possible from each other); and (\textit{iii})~we use a sigmoid rather than a softmax, which saves us from computing an additional sum in the denominator.

\subsubsection{Learning.}
During training, we optimise the following loss (Fig.~\ref{fig:overview}.A):
\begin{equation}
    \mathcal{L}_{training} = \mathcal{L}_{sim} + \lambda_{KL} \mathcal{L}_{KL} + \lambda_{var} \mathcal{L}_{var} + \lambda_{rep} \mathcal{L}_{rep},
\end{equation}
where we balance the hyperparameters $\{\lambda_{KL}, \lambda_{var}, \lambda_{rep}\}$ to avoid degenerate solutions such as empty features ($\mathcal{L}_{var}$), or keypoints at the image edges ($\mathcal{L}_{rep}$).

\subsection{Implementation details}
We demonstrate our three-fold regularisation loss on two existing methods: EquiTrack~\cite{billot_equitrack_2024} for rigid motion tracking, and KeyMorph~\cite{wang_keymorph_2023} for affine registration. For EquiTrack, we use SE(3)-equivariant layers of order 0, 1 and 2~\cite{e3nn}. For both EquiTrack and KeyMorph, the network $\Phi_\theta$ is implemented in PyTorch~\cite{paszke_pytorch_2019} as a 3D UNet~\cite{ronneberger_unet_2015} with 4 levels of 2 convolutions using 32 $3\!\times\!3\!\times\!3$ kernels, ReLU activations, and instance normalisation. We use a final per-channel softmax to separately normalise the $K=32$ output features. We train $\Phi_\theta$ with similar intensity and spatial augmentations as nnUNet~\cite{isensee_nnunet_2021}. After validation on dedicated data splits, we set $\{\lambda_{KL}, \lambda_{var}, \lambda_{rep}, \tau\} \!= \! \{1, 10^{-2}, 10^{-3}, 10^{-1}\}$.

\section{Experiments and results}

\begin{figure}[t]
\includegraphics[width=\textwidth]{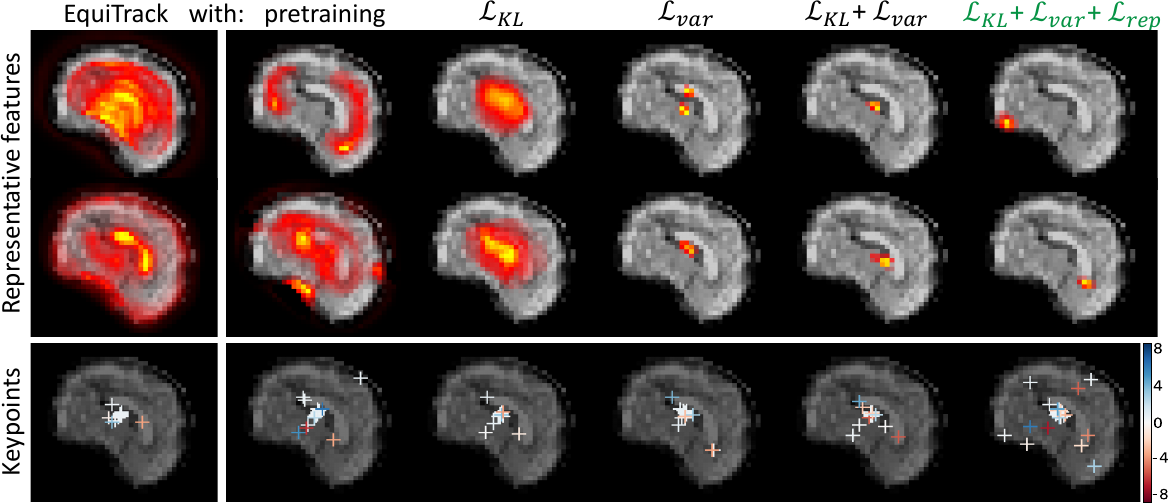}
\caption{Representative features and summary of all extracted keypoints obtained from an EPI foetal brain MRI with EquiTrack, trained with/without different regularisations. Keypoints are colour-coded by their distance to the visualised slice. When trained with our loss (green), EquiTrack yields interpretable, precise and diverse keypoints.} 
\label{fig:exp1}
\end{figure}

\begin{table}[t]
\caption{Average scores (with standard deviations in parentheses) obtained by EquiTrack and KeyMorph trained with/without different regularisations for rigid motion tracking in foetal MRI time series, as well as other competitors. Translation errors, spectral norms and average point distances are measured in voxels (3mm isotropic resolution). Best scores are in bold, and are found here to be always statistically significant at 1\% level with Bonferroni-corrected Wilcoxon signed-rank test.}
\setlength\tabcolsep{2.5pt}
\fontsize{8}{10}\selectfont
\begin{tabular}{|l|cc|ccc|}
\hline
Method & Rot. err. ($^\circ$) & Transl. err. & KL div. & Spectral norm & Point dist. \\
\hline
EquiTrack \cite{billot_equitrack_2024}                                                   & 5.1 (4.0)  & 0.8 (0.6) & 26.7 (2.1) & 46.3 (6.0) & 1.2 (0.5) \\
\hspace{0.05cm} $+$ pretraining from \cite{wang_keymorph_2023}                           & 4.6 (4.4)  & 0.7 (0.7) & 22.4 (3.8) & 43.1 (4.7) & 2.6 (1.3) \\
\hspace{0.05cm} $+\mathcal{L}_{KL}$                                                      & 3.7 (3.8)  & 0.8 (0.7) & \textbf{4.1 (1.4)} & 44.7 (6.3) & 1.5 (0.7) \\
\hspace{0.05cm} $+\mathcal{L}_{var}$                                                     & 3.6 (3.8)  & 0.7 (0.6) & 10.3 (3.4) & 5.4 (1.3)  & 1.9 (1.2) \\
\hspace{0.05cm} $+\mathcal{L}_{KL}\!+\!\mathcal{L}_{var}$                                & 2.4 (2.7)  & 0.5 (0.4) & 5.3 (1.7)  & 5.6 (1.2)  & 2.1 (1.4) \\
\hspace{0.05cm} $+\mathcal{L}_{KL}\!+\!\mathcal{L}_{var}\!+\!\mathcal{L}_{rep}\!$ (ours) & \textbf{1.6 (1.7)} & \textbf{0.3 (0.4)} & 4.7 (1.5) & 5.3 (1.2) & 4.1 (1.5) \\
\hline
KeyMorph \cite{wang_keymorph_2023}                                                       & 6.4 (6.4) & 0.8 (0.8) & 16.3 (4.5) & 36.1 (8.6) & 3.1 (1.2) \\
\hspace{0.05cm} $+\mathcal{L}_{KL}\!+\!\mathcal{L}_{var}\!+\!\mathcal{L}_{rep}\!$ (ours) & 2.7 (2.4) & 0.5 (0.5) & 4.4 (1.3)  & \textbf{4.9 (1.2)}  & \textbf{4.4 (1.6)} \\
\hline
ANTs \cite{avants_ants_2011} & 5.4 (4.2) & 0.6 (0.6) & - & - & - \\
\hline
DLIR \cite{devos_dlir_2019} & 9.9 (9.2) & 1.3 (1.4) & - & - & - \\
\hline
\end{tabular}
\label{tab:results_foetal}
\end{table}

\begin{figure}[t]
\includegraphics[width=\textwidth]{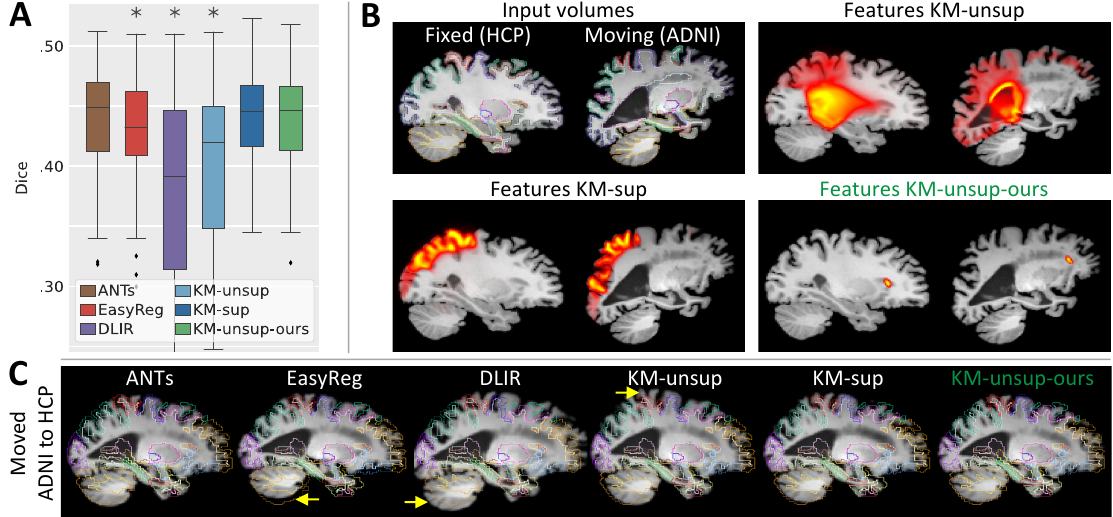}
\caption{(A)~Average Dice scores for ADNI to HCP brain MRI registration. Stars indicate statistical significance with KM-unsup-ours (1\% level, Wilcoxon signed-rank test). (B)~Example of HCP/ADNI scans with overlaid segmentations, and representative features from KeyMorph trained with different losses. (C)~Input ADNI~scan~affinely~regist-ered by all methods and segmentation of the fixed HCP scan. Arrows show large errors.}
\label{fig:exp2}
\end{figure}

\subsection{Rigid motion tracking in foetal MRI time series}
\label{sec:exp1}

\noindent\textbf{Task and dataset:} 
First, we test our method on a rigid motion tracking task, where our goal is to align all volumes in a time series to the first one. In this experiment, we use a private dataset of 55 3D foetal brain MRI time series (with approximately 28 scans each) from the Boston Children's Hospital. These are acquired on a 3T Skyra Siemens with a multi-slice single-shot gradient echo EPI sequence (TR=[5-8]ms, TE=[32-38]ms, $\alpha$=90\dg). We split time series into 30 (N=764 scans), 5 (N=152) and 20 (N=633) for training/validation/testing. Test scans come with manually extracted ground truth transforms. We emphasise that this data is challenging to analyse due to its low signal-to-noise ratio and low resolution (3mm isotropic).

We report rotation and translation errors, and three spatial metrics: KL divergence of the features (i.e., $\mathcal{L}_{KL}$); spectral norm (i.e., maximum eigenvalue) of their covariance matrix; and average distance between keypoints (Table~\ref{tab:results_foetal}).
\newline

\noindent\textbf{Results:}
We start with an ablation study to test the effect of the proposed regularisations on EquiTrack, which is the state of the art for foetal rigid motion tracking~\cite{billot_equitrack_2024}. While the original method yields relatively good alignments due to its SE(3)-equivariance (5.1\dg/0.8 vox. errors), all keypoints are centred and the features are hardly interpretable (Fig~\ref{fig:exp1}). In comparison, the pretraining strategy introduced in KeyMorph~\cite{wang_keymorph_2023} (see Section~\ref{sec:proposed losses}) slightly increases the mean point distance by 1.4 voxels, which leads to a 0.5\dg~improvement. Then,~gradually adding $\mathcal{L}_{KL}$ and $\mathcal{L}_{var}$ further improves accuracy (Table~\ref{tab:results_foetal}) by encouraging more explainable and precise features. Finally, due to the explicit loss for keypoint diversity (Fig~\ref{fig:exp1}), the fully regularised EquiTrack achieves highly accurate alignments, with improvements over the original method of 3.5\dg and 0.5 voxels. Remarkably, this boost in accuracy comes at no supervision cost, and only requires slightly longer training times (+10\%, 60h total on a A6000 Nvidia GPU).

We then show that our loss also applies to other architectures by testing it on KeyMorph~\cite{wang_keymorph_2023}, that we adapt to the rigid case by using the closed-form algorithm from~\cite{kabsch_svd_1976}. Here, our loss yields an improvement of 3.7\dg and 0.3 voxels over the original KeyMorph (Table~\ref{tab:results_foetal}). Notably, this is higher than the architecture-based improvement achieved by replacing the CNN in KeyMorph by an equivariant CNN (i.e., vanilla EquiTrack), thus showing the effectiveness of our approach.

Finally, we compare against two strategies for rigid registration: the ubiquitous optimisation framework ANTs~\cite{avants_ants_2011}, and the deep learning regression-based strategy DLIR~\cite{devos_dlir_2019}. Table~\ref{tab:results_foetal} shows that our fully regularised EquiTrack beats ANTs by 3.8\dg and 0.3 voxels (with faster runtimes by 2 orders of magnitude) and considerably outperforms DLIR by 8.3\dg, while being much more interpretable.

\subsection{Affine brain MRI registration}
\label{sec:exp2}

\noindent\textbf{Task and dataset:} 
We then assess our approach in the challenging scenario of affine registration between brain MRIs from different populations (with applications in population atlas building~\cite{chen_survey_2025}). Here we register 200 ageing patients from ADNI~\cite{jack_adni_2008} onto 200 young and healthy subjects from HCP~\cite{vanessen_hcp_2012}. All volumes are 1mm isotropic T1-weighted scans. We use splits of~40/10/50\% for training/validation/testing. Since ground truth transforms are not available, we evaluate performance by segmenting all scans with SynthSeg for 98 regions~\cite{billot_synthseg_2023} and by computing Dice scores between the fixed and moved segmentations.

Since EquiTrack cannot handle inter-subject registration~\cite{billot_equitrack_2024}, we only test our loss on KeyMorph, for which we train three variants: original KeyMorph (KM-unsup), regularised KeyMorph (KM-unsup-ours), and a supervised version (KM-sup) optimised with a Dice loss~\cite{wang_keymorph_2023}. We also include affine versions of ANTs, DLIR, and EasyReg~\cite{easyreg_iglesias_2023}, which is a supervised keypoint-based method, where keypoints are the centres of mass of the 98 regions segmented by SynthSeg. 

\noindent\textbf{Results:}
First, our method KM-unsup-ours yields similar scores to the state-of-the-art ANTs (Fig.~\ref{fig:exp2}.C) while running much faster, which enables time-sensitive applications. Then, KM-unsup-ours largely outperforms the other deep learning unsupervised approaches DLIR and KM-unsup by 5.6 and 3.1 Dice points, respectively (Fig.~\ref{fig:exp2}.A). Further, our method also beats EasyReg by 1.5 Dice points, which is due to the fact that region centres used in EasyReg might~not be reliable keypoints since their relative location may vary across populations (e.g.,``shifted'' centres of mass for small vs. enlarged ventricles). Crucially, our unsupervised loss bridges the performance gap with the supervised KM-sup (no statistical difference). This is explained by our proposed regularisation, which enables KM-unsup-ours to extract precise and anatomically meaningful keypoints that are consistent across subjects (Fig.~\ref{fig:exp2}.B), without requiring any supervision.

\section{Conclusion}

We have presented a novel principled regularisation method for registration based on unsupervised keypoint detection. Our strategy greatly increases the interpretability of unsupervised strategies and leads to consistent improvements for two state-of-the-art methods across different tasks and datasets. Specifically, the impact of the proposed loss is comparable to that of major architectural changes (Section~\ref{sec:exp1}), and bridges the gap with supervised methods (Section~\ref{sec:exp2}). Future work will tackle cross-modality registration by combining our loss with supervision from segmentations. Another direction is to dynamically learn the number of keypoints to further increase diversity. Overall, our work opens new perspectives for the interpretability of modern deep learning registration models.


\begin{credits}
\subsubsection{\ackname} This work has been funded by NIH NIBIB 1R01EB036945, NIH NICHD 1R01HD114338, NIH NIBIB
1R01EB032708, MIT Jameel Clinic, MIT CSAIL-Wistron Program. Further support has come from the French government, through the 3IA Cote d’Azur Investments in the project managed by the National Research Agency (ANR) with the reference number ANR-23-IACL-0001, 
\subsubsection{\discintname} The authors have no competing interests to declare.
\end{credits}


\begin{thebibliography}{10}
\providecommand{\url}[1]{\texttt{#1}}
\providecommand{\urlprefix}{URL }
\providecommand{\doi}[1]{https://doi.org/#1}

\bibitem{avants_symmetric_2008}
Avants, B., Epstein, C., Grossman, M., Gee, J.: Symmetric diffeomorphic image registration with cross-correlation: evaluating automated labeling of elderly and neurodegenerative brain. Medical Image Analysis  \textbf{12},  26--41 (2008)

\bibitem{avants_ants_2011}
Avants, B., Tustison, N., Song, G., Cook, P., Klein, A., Gee, J.: A reproducible evaluation of ants similarity metric performance in brain image registration. Neuroimage  \textbf{54},  2033--2044 (2011)

\bibitem{balakrishnan_voxelmorph_2019}
Balakrishnan, G., Zhao, A., Sabuncu, M., Guttag, J., Dalca, A.: Voxelmorph: A learning framework for deformable medical image registration. IEEE transactions on Medical Imaging  \textbf{38},  788--800 (2019)

\bibitem{billot_equitrack_2024}
Billot, B., Dey, N., Moyer, D., Hoffmann, M., {Abaci Turk}, E., Gagoski, B., Grant, E., Golland, P.: {SE}(3)-equivariant and noise-invariant {3D} rigid motion tracking in brain {MRI}. IEEE transactions on Medical Imaging  \textbf{43},  4029--4040 (2024)

\bibitem{billot_synthseg_2023}
Billot, B., Greve, D., Puonti, O., Thielscher, A., {Van Leemput}, K., Fischl, B., Dalca, A., Iglesias, J.E.: Synthseg: Segmentation of brain {MRI} scans of any contrast and resolution without retraining. Medical Image Analysis  \textbf{86} (2023)

\bibitem{chatrasingh_svd-scaling_2023}
Chatrasingh, M., Wiratkapun, C., Suthakorn, J.: A generalized closed-form solution for {3D} registration of two-point sets under isotropic and anisotropic scaling. Results in Physics  \textbf{51} (2023)

\bibitem{chen_survey_2025}
Chen, J., Liu, Y., Wei, S., Bian, Z., Subramanian, S., Carass, A., Prince, J., Du, Y.: A survey on deep learning in medical image registration: New technologies, uncertainty, evaluation metrics, and beyond. Medical Image Analysis  \textbf{100} (2025)

\bibitem{chui_point_2003}
Chui, H., Rangarajan, A.: A new point matching algorithm for non-rigid registration. Computer Vision and Image Understanding  \textbf{89},  114--141 (2003)

\bibitem{dalca_unsupervised_2019}
Dalca, A., Balakrishnan, G., Guttag, J., Sabuncu, M.: Unsupervised learning of probabilistic diffeomorphic registration for images and surfaces. Medical Image Analysis  \textbf{57},  226--236 (2019)

\bibitem{devos_dlir_2019}
{de Vos}, B., Berendsen, F., Viergever, M., Sokooti, H., Staring, M., Išgum, I.: A deep learning framework for unsupervised affine and deformable image registration. Medical Image Analysis  \textbf{52},  128--143 (2019)

\bibitem{eppenhof_pulmonary_2018}
Eppenhof, K., Pluim, J.: Pulmonary ct registration through supervised learning with convolutional neural networks. IEEE transactions on Medical Imaging  \textbf{38},  1097--1105 (2018)

\bibitem{vanessen_hcp_2012}
Essen, V., et~al.: The human connectome project: a data acquisition perspective. Neuroimage  \textbf{62},  222--231 (2012)

\bibitem{evan_keymorph_2022}
Evan, Y., Wang, A., Dalca, A., Sabuncu, M.: {KeyMorph}: Robust multi-modal affine registration via unsupervised keypoint detection. MIDL  (2022)

\bibitem{fan_birnet_2019}
Fan, J., Cao, X., Yap, P., Shen, D.: {BIRNet}: Brain image registration using dual-supervised fully convolutional networks. Medical Image Analysis  \textbf{54},  193--206 (2019)

\bibitem{fang_var_2023}
Fang, Z., Delingette, H., Ayache, N.: Anatomical landmark detection for initializing {US} and {MR} image registration. International Workshop on Advances in Simplifying Medical Ultrasound pp. 165--174 (2023)

\bibitem{e3nn}
Geiger, M., Smidt, T.: e3nn: Euclidean neural networks (2022), \url{https://e3nn.org}

\bibitem{holland_longitudinal_2011}
Holland, D., Dale, A.: Nonlinear registration of longitudinal images and measurement of change in regions of interest. Medical Image Analysis  \textbf{15},  489--497 (2011)

\bibitem{horn_svd_1987}
Horn, B.: Closed-form solution of absolute orientation using unit quaternions. Journal of the Optical Society  \textbf{4},  629--642 (1987)

\bibitem{hu_weakly_2018}
Hu, Y., Modat, M., Gibson, E., Ghavami, N., Bonmati, E., Moore, C., Emberton, M., Noble, A., Barratt, D., Vercauteren, T.: Weakly-supervised convolutional neural networks for multimodal image registration. Medical Image Analysis  (2018)

\bibitem{easyreg_iglesias_2023}
Iglesias, J.E.: Weakly-supervised convolutional neural networks for multimodal image registration. Scientific Reports  \textbf{13},  1--15 (2023)

\bibitem{isensee_nnunet_2021}
Isensee, F., Jaeger, P., Kohl, S., Petersen, J., Maier-Hein, K.: {nnU-Net}: a self-configuring method for deep learning-based biomedical image segmentation. Nature methods  \textbf{18},  203--211 (2021)

\bibitem{jack_adni_2008}
Jack~Jr., C.R., et~al.: The alzheimer's disease neuroimaging initiative ({ADNI}): {MRI} methods. Journal of Magnetic Resonance Imaging  \textbf{27},  685--691 (2008)

\bibitem{joshi_atlas_2004}
Joshi, S., Davis, B., Jomier, M., Gerig, G.: Unbiased diffeomorphic atlas construction for computational anatomy. NeuroImage  \textbf{23},  151--160 (2004)

\bibitem{kabsch_svd_1976}
Kabsch, W.: A solution for the best rotation to relate two sets of vectors. Acta Crystallographica  \textbf{32},  922--923 (1976)

\bibitem{lee_image_2019}
Lee, M., Oktay, O., Schuh, A., Schaap, M., Glocker, B.: Image-and-spatial transformer networks for structure-guided image registration. MICCAI pp. 37--45 (2019)

\bibitem{liu_anatomy_2021}
Liu, L., Wolterink, J., Brune, C., Veldhuis, R.: Anatomy-aided deep learning for medical image segmentation: a review. Physics in Medicine \& Biology  \textbf{66} (2021)

\bibitem{lorenzi_lcc_2013}
Lorenzi, M., Ayache, N., Frisoni, G., Pennec, X.: Lcc-demons: a robust and accurate symmetric diffeomorphic registration algorithm. NeuroImage  \textbf{81},  470--483 (2013)

\bibitem{modat_niftyreg_2014}
Modat, M., Cash, D., Daga, P., Winston, G., Duncan, J., Ourselin, S.: Global image registration using a symmetric block-matching approach. Journal of Medical Imaging  \textbf{1},  21--32 (2014)

\bibitem{salehi_3dposenet_2019}
Mohseni~Salehi, S., Khan, S., Erdogmus, D., Gholipour, A.: Real-time deep pose estimation with geodesic loss for image-to-template rigid registration. IEEE transactions on Medical Imaging  \textbf{38},  470--481 (2019)

\bibitem{moyer_equivariant_2021}
Moyer, D., {Abaci Turk}, E., Grant, E., Wells, W., Golland, P.: Equivariant filters for efficient tracking in {3D} imaging. MICCAI pp. 193--202 (2021)

\bibitem{nibali_dnst_2018}
Nibali, A., He, Z., Morgan, S., Prendergast, L.: Numerical coordinate regression with convolutional neural networks. arXiv preprint arXiv:1801.07372  (2018)

\bibitem{paszke_pytorch_2019}
Paszke, A., et~al.: Pytorch: An imperative style, high-performance deep learning library. Advances in neural information processing systems  \textbf{32} (2019)

\bibitem{pennec_landmark_1997}
Pennec, X., Thirion, J.P.: A framework for uncertainty and validation of {3D} registration methods based on points and frames. International Journal of Computer Vision  \textbf{25},  203--229 (1997)

\bibitem{ronneberger_unet_2015}
Ronneberger, O., Fischer, P., Brox, T.: U-{Net}: Convolutional networks for biomedical image segmentation. MICCAI pp. 234--241 (2015)

\bibitem{rueckert_bsplines_1999}
Rueckert, D., Sonoda, L., Hayes, C., Hill, D., Leach, M., Hawkes, D.: Nonrigid registration using free-form deformations: application to breast {MR} images. IEEE transactions on Medical Imaging  \textbf{18},  712--721 (1999)

\bibitem{tuytelaars_landmarks_2008}
Tuytelaars, T., Mikolajczyk, K.: Local invariant feature detectors: a survey. Foundations and trends in computer graphics and vision  \textbf{3},  177--280 (2008)

\bibitem{uzunova_training_2017}
Uzunova, H., Wilms, M., Handels, H., Ehrhardt, J.: Training {CNN}s for image registration from few samples with model-based data augmentation. MICCAI pp. 223--231 (2017)

\bibitem{wang_keymorph_2023}
Wang, A., Yu, E., Dalca, A., Sabuncu, M.: A robust and interpretable deep learning framework for multi-modal registration via keypoints. Medical Image Analysis  \textbf{90} (2023)

\bibitem{wang_understanding_2020}
Wang, T., Isola, P.: Understanding contrastive representation learning through alignment and uniformity on the hypersphere. International Conference on Machine Learning pp. 929--939 (2020)

\end{thebibliography}

\end{document}